\documentclass[10pt,letterpaper]{article}
\usepackage[top=0.85in,left=1in,footskip=0.75in,marginparwidth=2in]{geometry}
\pdfoutput=1

\usepackage[utf8]{inputenc}

\usepackage{cite}
\usepackage{nameref,hyperref}

\usepackage{microtype}
\DisableLigatures[f]{encoding = *, family = * }

\usepackage{booktabs, float}
\usepackage{algorithm}
\usepackage{algpseudocode}
\usepackage{array}
\usepackage{threeparttable} 
\renewcommand{\thetable}{\arabic{table}} 

\usepackage{ragged2e} 
\usepackage{indentfirst}
\justifying
\usepackage{changepage}

\usepackage[aboveskip=1pt,labelfont=bf,labelsep=period,singlelinecheck=off]{caption}

\makeatletter
\renewcommand{\@biblabel}[1]{\quad#1.}
\makeatother

\usepackage{lastpage,fancyhdr,graphicx}
\usepackage{epstopdf}
\fancyheadoffset[L]{2.25in}
\fancyfootoffset[L]{2.25in}

\usepackage{color}

\definecolor{Gray}{gray}{.25}

\usepackage{graphicx}

\usepackage{sidecap}

\usepackage{wrapfig}
\usepackage[pscoord]{eso-pic}
\usepackage[fulladjust]{marginnote}
\reversemarginpar

\usepackage{hyperref}  
\usepackage{xurl}      
\usepackage{breakurl}  

\makeatletter
\g@addto@macro\UrlBreaks{\do\/\do-\do\.\do\_\do\%\do\&\do\#\do\?}
\makeatother
\begin{document}
\vspace*{0.35in}

\begin{flushleft}
{\Large
\textbf\newline{Text Chunking for Document Classification for Urban Systems Management using Large Language Models}
}
\newline
\\
Joshua Rodriguez\textsuperscript{1†},
Om Sanan\textsuperscript{2†},
Guillermo Vizarreta-Luna\textsuperscript{1},
Steven A. Conrad\textsuperscript{1*}
\\
\bigskip
\bf{1} Department of Systems Engineering, Colorado State University, Fort Collins, CO, USA
\\
\bf{2} Scarsdale High School, Scardsale, NY, USA
\\
\bigskip
* Corresponding author; email: steve.conrad@colostate.edu
\\
† Equal Contribution
\end{flushleft}

\section*{Abstract}
Urban systems are managed using complex textual documentation that need coding and analysis to set requirements and evaluate built environment performance. This paper contributes to the study of applying large language models (LLM) to qualitative coding activities to reduce resource requirements while maintaining comparable reliability to humans. Qualitative coding and assessment face challenges like resource limitations and bias, accuracy, and consistency between human evaluators. Here we report the application of LLMs to deductively code 10 case documents on the presence of 17 digital twin characteristics for the management of urban systems. We utilize two prompting methods to compare the semantic processing of LLMs with human coding efforts: whole text analysis and text chunk analysis using OpenAI’s GPT-4o, GPT-4o-mini, and o1-mini models. We found similar trends of internal variability between methods and results indicate that LLMs may perform on par with human coders when initialized with specific deductive coding contexts. GPT-4o, o1-mini and GPT-4o-mini showed significant agreement with human raters when employed using a chunking method. The application of both GPT-4o and GPT-4o-mini as an additional rater with three manual raters showed statistically significant agreement across all raters, indicating that the analysis of textual documents is benefited by LLMs. Our findings reveal nuanced sub-themes of LLM application suggesting LLMs follow human memory coding processes where whole-text analysis may introduce multiple meanings. The novel contributions of this paper lie in assessing the performance of OpenAI GPT models and introduces the chunk-based prompting approach, which addresses context aggregation biases by preserving localized context.


\section{Introduction}
The management of Urban Systems – the interconnection of built environment, the natural environment, and society – involves complex textual documentation that requires persistent review to derive requirements and performance standards. Architects and planners review codes and regulations to ensure construction projects align with current standards. Water managers review environmental protection rules and scientific reports to ensure water quality, and policymakers review public consultation documents, policy proposals, and research studies to assess urban development projects. Each of these roles necessitates humans to read, digest, and semantically process hundreds of thousands of words to deduce relationships and presence of content for a specific theme or premise. Accurate analysis of these documents directly influences the execution of critical operational processes in urban systems and can impact human welfare. Missteps in the coding process could lead to omissions or inaccuracies, potentially skewing broader urban system designs and policy implementations. Substantially large data sets may preclude analysis and remain unexamined. 

Traditional methods for coding documentation, such as human deductive coding, presents resource challenges and outcomes varies by human coder. Human coders present biases, variations in accuracy and consistency \cite{gonzalez_canche_latent_2023}. Manual textual analysis by humans is prone to notable inconsistencies and errors, especially when managing large datasets or working with multiple coders \cite{gonzalez_canche_latent_2023}. Humans are good at understanding complex meanings and context in texts \cite{hillard_computer-assisted_2008}, but the manual coding of extensive data sets can be challenging due to its time and cost-intensive nature \cite{chang_accelerating_2021,hillard_computer-assisted_2008} and consistent application of rules across individuals \cite{baumer_comparing_2017}. Artificial Intelligence (AI) presents the opportunity to enhance processes through human-AI teaming in domain-agnostic tasks \cite{klami_virtual_2024}. The recent introduction of Large Language Models (LLMs) presents an opportunity for Urban Systems to reduce the burdens of documentation management. 

Within limits, pretrained LLMs show potential to augment deductive coding. Other forms of AI have been shown for unstructured data tasks, such as content-based image retrieval \cite{shim_content-based_2023, marathe_semi-automated_2018,desmond_semi-automated_2021}. For instance, LLMs can be used to for classification and information retrieval from unstructured, text-based files \cite{devlin_bert_2019,radford_improving_2018}. LLMs, however, frequently focus on next-word prediction leading to propagated biases and incoherent texts \cite{bender_dangers_2021, radford_language_2019}. In addition, LLMs are subject to hallucinations or fabricating information due to false or inadequate training data and lacking knowledge recall processes \cite{huang_survey_2023}. To combat this, Ouyang et al. \cite{ouyang_training_2022} utilized human input for fine-tuning LLM responses using reward modelling and reinforcement learning to align LLM objectives with operator goals. Raczyński et al. \cite{gal_problem_2023} addressed issues of natural language model coherence through the application of a transformer to increase explainability of the outputs of language models. In addition, it has been shown that general language models perform better at extracting information from text-based data than off-the-shelf language models when provided with prior knowledge enhancement or a more technical training set \cite{gal_diversified_2023, lee_biobert_2020}. These studies indicate the need for adequate prompting before language analysis tasks.

The application of LLMs in qualitative research requires rigorous processes. Tai et al. \cite{tai_examination_2024} stresses the importance of establishing robust analytical practices and protocols. They note that prompts designed to align with a codebook can effectively replicate traditional document classification methods. Expanding on this, Xiao et al. \cite{xiao_supporting_2023} explore using GPT-3 for deductive coding in qualitative research by integrating LLMs with expert-developed codebooks containing document classification categories. Their methodology analyzes children’s curiosity-driven questions using codebook-centered and example-centered prompts, testing variations in prompt complexity. They find GPT-3 achieved substantial agreement with experts for question complexity and fair agreement for syntactic structure, with codebook-centered prompts performing best. Moreover, Chew et al. \cite{chew_llm-assisted_2023} investigate GPT-3.5 for deductive coding through their LLM-Assisted Content Analysis framework which evaluates LLM performance in traditional qualitative workflows, emphasizing tasks such as co-developing codebooks, refining prompts, and testing reliability in line with human coding benchmarks. They conclude that LLMs achieved accuracy comparable to human coders but emphasized their role as tools to support, not replace, researchers. Prescott et al. \cite{prescott_comparing_2024} similarly test both Open AI’s ChatGPT 3.5 and Google’s Bard models in deductive coding applications for SMS messages, finding poor consistency with humans. 

In this paper we explore the application of LLMs to address the resource and consistency challenges of humans coding complex documents. We utilize the context of managing urban systems as a test case. We examine the premise that LLMs could reduce the burden of coding while maintaining reliability comparable to human coders. Specifically, we investigate the use of OpenAI’s GPT-4o, GPT-4o-mini, and o1-mini models to deductively code digital twin characteristics from literature on urban water systems. We examine two semantic processing methods: whole text analysis and chunking text into smaller sections and compare the performance to human evaluators. 

While previous research has provided insights on using LLMs for deductive coding, our study takes a novel approach by comparing whole text and text chunk analysis using the three distinct OpenAI LLMs. In doing so this research explores how the volume of data impacts LLM accuracy and effectiveness. Another key contribution of our research is addressing context aggregation biases through a fixed-size text chunk approach. Fixed-sized text chunking may preserve localized and sectional themes that are overlooked when treating entire documents as single units. This paper explores the performance of text chunking, illustrating that its application is appropriate for managing the size of documents analyzed by LLMs, providing insights that expose nuances in meanings that whole text analysis might miss.

\section{Materials and Methods}
\subsection{Deductive Coding}
The deductive coding approach utilized a detailed codebook of 17 digital twin (DT) characteristics adapted from \cite{jones_characterising_2020}, with expanded definitions providing clear context and examples for each parameter (a subset is shown in Table \ref{tab1}). Each dimension was marked as True or False for a paper, with True indicating that the paper discussed the dimension of interest as defined. Ten articles discussing the development of digital twins were selected for coding. Deductive coding was performed manually by a team of three non-expert researchers following the methods outlined in \cite{elo_qualitative_2008} to determine if the papers discussed each of the DT characteristics. This manual coding served as the reference point for assessing the performance of the LLMs. Deductive coding by LLMs was then performed using the same codebook given to humans. 

\begin{table}[!ht]
    \centering
    \caption{Subset of the codebook for the 17 dimensions coded in the paper. A full codebook is available in the supplemental materials.}
    \label{tab1}

    \renewcommand{\arraystretch}{1.8} 

    \begin{tabular}{p{3cm} p{12cm}} 
        \toprule
        \textbf{Dimension} & \textbf{Definition} \\
        \midrule
        \textbf{Physical Entity or Process} & A tangible object or a real-world process within a physical system, such as a piece of equipment, a component, or an operational process in an industrial setting such as a manufacturing operation. \\ 
        \textbf{Fidelity} & Degree of accuracy and completeness with which a digital twin replicates the physical counterpart of the system, related to data, behavior, physical structure, etc. Examples include “comprehensive physical and functional description” or fully mirroring the characteristics and functionality of the physical entity.\\
        \textbf{Use-Cases} & The applications of the Digital Twin, e.g., decision support, simulation, forecasting. \\
        \bottomrule
    \end{tabular}
\end{table}

\subsection{LLM Utilization}
Two LLM prompting approaches were used for comparison, utilizing three of OpenAI's latest models: `gpt-4o`, `gpt-4o-mini`, and `o1-mini `. Notably, `o1-mini` employs a recursive thought chain mechanism, offering potentially enhanced reasoning capabilities. Text was extracted from PDF files of the publications using PyPDF2, with multiple verification processes and Optical Character Recognition technology from Python’s PyTesseract and pdf2image libraries to enable effective analysis of scanned documents and images. Extracted texts underwent preprocessing to remove unwanted characters, correct hyphenations, and fix erroneous newline insertions. The documents were divided into fixed-size 500-word chunks using a chunking mechanism to facilitate manageable input sizes for the LLMs. Each chunk was analyzed individually for the presence of each DT characteristic. The OpenAI Python API was used to interface with the LLMs, resetting the LLM instance between each prompt execution to ensure independence of responses. Algorithm \ref{alg1} and Algorithm \ref{alg2} illustrate the methods used for each approach. Method 1 (the “Whole Paper Approach”) passes the entire text to the LLM, along with the associated prompt for each characteristic. Method 2 (the “Chunking Approach”) processes the document in 500-word chunks, passing each chunk to the LLM for individual deductive coding analysis. 
In this study, prompt engineering was not explored, and the LLM was given the same codebook as the manual raters. The structure of the prompt passed to the LLM can likely influence the results of the response, although this was deemed out of scope for this research. The following prompt was used for all models: 

\textit{“Explain whether the parameter '{parameter}' is mentioned/directly talked about in the following text and provide evidence from the text. If it does, briefly explain how (3-5 sentences with ~2 pieces of evidence); if it does not match, briefly explain why the paper does not focus on it (1 sentence). Note that '{parameter}' is defined as '{definition}'." }

\begin{minipage}{0.48\textwidth} 
    \begin{algorithm}[H]
        \caption{Whole Paper analysis of documents, taking as input StudySet, Codebook}
        \label{alg1}
        \begin{algorithmic}[1]
            \For{Text $\in$ StudySet}
                \For{Dim $\in$ Codebook}
                    \State PASS Text to LLM
                    \State PROMPT LLM to determine if Dim is in Text
                    \If{LLM = TRUE}
                        \State Dim $\gets$ TRUE
                    \EndIf
                    \State OutputTable(Text, Dim) $\gets$ Dim
                \EndFor
                \State \Return OutputTable(Text)
            \EndFor
        \end{algorithmic}
    \end{algorithm}
\end{minipage}
\hfill
\begin{minipage}{0.48\textwidth} 
    \begin{algorithm}[H]
        \caption{Fixed-Size Chunk analysis of documents, taking as input StudySet, Codebook}
        \label{alg2}
        \begin{algorithmic} [1]
            \For{Text $\in$ StudySet}
                \State Divide Text into TextChunks
                \For{Item $\in$ TextChunks} 
                    \For{Dim $\in$ Codebook}
                        \State PASS Text to LLM
                        \State PROMPT LLM to determine if Dim is in Text
                        \If{LLM = TRUE}
                            \State Dim $\gets$ TRUE
                        \EndIf
                        \State OutputTable(Text, Dim) $\gets$ Dim
                    \EndFor
                \EndFor
                \State \Return OutputTable(Text)
            \EndFor
        \end{algorithmic}
    \end{algorithm}
    \vspace{5pt}
\end{minipage}

In this paper, a “consensus approach” was explored where the mode of a dimension for a given paper across all fifteen iterations (either True or False) was selected as the output of the total LLM analysis, essentially introducing the LLM as an additional individual rater. The consensus approach was examined for its accuracy as an individual rater (using the consensus of the three manual raters as a gold standard for correct classifications) and for its applications as an additional rater in conjunction with the manual team, an exploration of human-AI teaming for document analysis. Sample outputs from the models are shown in Table \ref{tab:sample} in the appendix, including a positive and negative result for each. The results from the chunking approach were a stitching of these outputs, with one response for every 500 words in the article rounded up.

Return results would then be classified as true if a minimum of one response indicated the presence of a dimension. Following the classification by the LLM, a keyword search was employed to mark outputs of results. The first 510 results from the whole paper approach (170 results for each model) were used to identify the keywords and phrases indicating a true result. Out of these results, 1.7\% were incorrectly classified using this approach, with a near 50-50 split of false positives (FP) and false negatives (FN) (5 FP – 4 FN), indicating a negligible level of error from the key phrase classification. The following key phrases were used to mark responses from the LLM as true: 

\textit{\{yes; clearly stated; the text does mention; the text does discuss; the paper mentions the parameter; indirectly mentioned; is explicitly mentioned; indeed; does talk; is discussed; is referenced; is mentioned; implicit; does address\}}

The performance of the deductive coding approach was assessed using interrater agreement, accuracy, precision, and recall. Measures of accuracy, precision, and recall were calculated according to \cite{witten_credibility_2011}. The Fleiss' Kappa (K) value is used to measure agreement between raters when the raters are different for each subject, but the volume of raters is the same \cite{fleiss_measuring_1971}. Fleiss’ Kappa values were calculated using the `kappam.fleiss` function from the `irr` v0.84.1 package for R 2023.12.1+402 \cite{gamer_irr_2010}. The performance of the LLMs were also evaluated using the internal variance which was measured by calculating the percentage of iterations which adhered to the consensus result. 
Percent agreement is a measure of the agreement amongst raters, representing the proportion of responses between the raters which align with one another and the number of non-erroneous results \cite{mchugh_interrater_2012}. As a binary scale is used in this research, the percent agreement is taken as the proportion of responses which align with the mode across the raters. The percent agreement metric is limited as it does not account for the possibility of chance agreements \cite{mchugh_interrater_2012}. In addition, percent agreement is less effective when employed in classification systems with non-binary or hierarchical levels as the difference in ratings may have variable magnitudes \cite{graham_measuring_2014}. When using percent agreement as an interrater reliability metric, higher standards must be achieved to address the assumption that all agreements are not driven by chance, thus it is recommended by Graham et al. \cite{graham_measuring_2014} that a target percent agreement of 90\% should be achieved for the classifications to be considered strong, especially when used with adjacent categories. Fleiss’ kappa coefficient is known to be subject to paradoxical behavior, where the kappa may be underestimated, even as percent agreement is high \cite{falotico_fleiss_2015}. For the Fleiss kappa coefficient, \cite{fleiss_statistical_2003} recommend a range of 0.40-0.75 for fair agreement beyond chance, while at scores below 0.40 agreement is strongly driven by chance agreements. 

\section{Results}

Table \ref{tab:performance_comparison} presents our findings across the measures of variance, accuracy, precision, and recall. Variance is measured by internal agreement across iterations. Success rate (herein called Accuracy), precision, and recall are metrics to evaluate data mining techniques as proposed by Witten and coworkers \cite{witten_credibility_2011}. As defined by \cite{witten_credibility_2011}, recall is the rate of relevant retrievals to all relevant items while precision is the rate of relevant retrievals to all retrievals. Accuracy is the total relevant retrievals to the entire body of mined data. In the subsequent sections we discuss each measure in detail.  
\begin{table}[!ht]
    \centering
    \caption{Performance comparison between treatment (models and prompting approaches) representing the mean value across the 10 papers analyzed across 15 iterations as compared to the control (human raters)}
    \label{tab:performance_comparison}
    \resizebox{\textwidth}{!}{ 
    \begin{threeparttable}
    \begin{tabular}{lcccccc}
        \toprule
        \textbf{Measure} & \multicolumn{2}{c}{\textbf{GPT-4o}} & \multicolumn{2}{c}{\textbf{GPT-4o-mini}} & \multicolumn{2}{c}{\textbf{o1-mini}} \\
        \cmidrule(lr){2-3} \cmidrule(lr){4-5} \cmidrule(lr){6-7}
        & Whole Paper & Chunking & Whole Paper & Chunking & Whole Paper & Chunking \\
        \midrule
        Iteration Percent Agreement$^a$ *** & 73.55\% & 87.61\%\textsuperscript{+++} & 78.88\% & 80.35\% & 64.47\% & 89.89\%\textsuperscript{+++} \\
        Accuracy ** & 74.12\% & 87.65\%\textsuperscript{++} & 77.65\% & 84.12\% & 72.35\% & 87.06\%\textsuperscript{++} \\
        Precision & 91.68\% & 90.63\% & 90.17\% & 93.64\% & 93.85\% & 89.90\% \\
        Recall *** & 76.58\% & 95.09\%\textsuperscript{+++} & 82.61\% & 86.52\% & 72.18\% & 94.91\%\textsuperscript{+++} \\
        \bottomrule
    \end{tabular}
    \begin{tablenotes}
        \small
        \item $^a$ Represents internal percent agreement without comparison to human raters.
        \item *, **, *** Represents Kruskal-Wallis significance difference at 10\%, 5\%, 1\% between treatments.
        \item +, ++, +++ Represents pairwise Mann-Whitney comparison significance at 10\%, 5\%, 1\% between prompting approaches of the same model.
    \end{tablenotes}
    \end{threeparttable}
    }
\end{table}

\subsection{LLM Internal Variance}
After 15 iterations and across all authors, utilizing the chunking approach showed higher average internal agreement across all models, indicating that it had higher consistency during repeated prompting compared to the whole paper approach. The o1-mini model’s chunking approach had the highest internal agreement at 89.9\% compared to the whole paper approach of 64.5\%. GPT-4o and GPT-4o-mini respectively showed average internal agreements across all papers as 87.6\% and 80.4\% for the chunking approach and 73.5\% and 78.9\% for the whole paper approach. 
Nonparametric statistics were employed in this analysis because of the non-normality of the o1-mini chunking approach internal agreements. Using the final internal agreement percentage by paper, a Kruskal-Wallace test shows a p-value of 1.71x10-5, indicating that there is a statistically significant difference between the means of some of the groups. Pairwise Mann-Whitney testing showed that the chunking approach for o1-mini had statistically significant more internal agreement than all whole paper methods and the GPT-4o-mini chunking approach. We found that o1-mini was the only model that showed significant improvements in internal agreement when moving from the whole paper approach to the chunking approach. The o1-mini whole paper approach was significantly less consistent than all models employing the chunking approach. While the averaged internal agreement for the chunking approach is greater than the whole paper, there is no statistically significant difference in internal agreement between the models and prompting approaches. This indicates that o1-mini is significantly impacted by the volume of data passed compared to the other models, and that with a smaller context it can perform better than the other two models. Neither GPT-4o or GPT-4o-mini when chunked shows significant variance from the whole paper approach or from the other models.  
When performing single-sample Wilcoxon Signed-Rank Test, all whole paper approaches and the GPT-4o-mini chunking approach show significant difference from a median internal agreement value of 90\% while the chunking approaches for GPT-4o and o1-mini show no significant difference (p \textgreater .10). As Graham et al. \cite{graham_measuring_2014} and Stemler \cite{stemler_comparison_2004} point out, this indicates that the agreements between the whole paper approaches and the manual classifications may be driven by chance, even though their accuracy is not significantly different from the chunking approaches.

\begin{figure}[H]
    \centering
    \includegraphics[width=.65\linewidth]{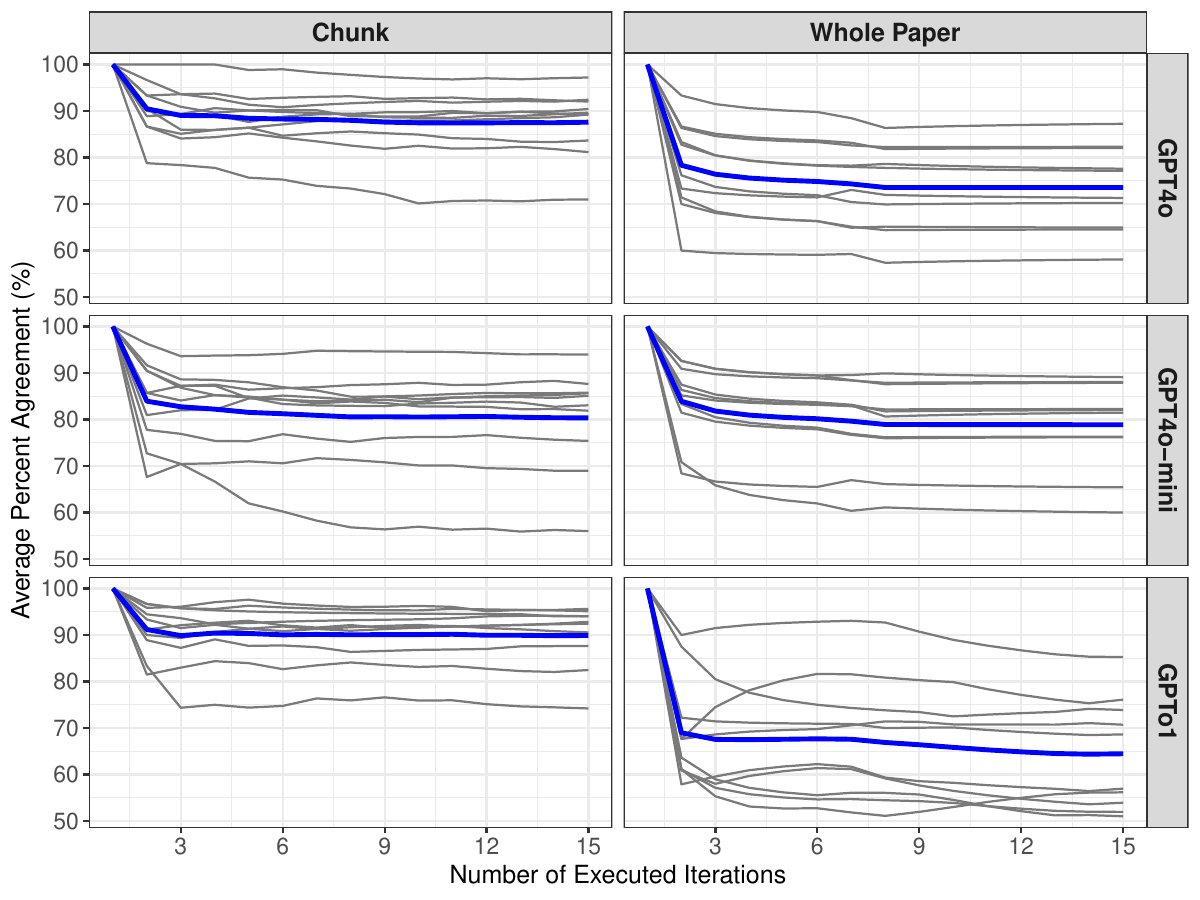}
    \caption{Internal Agreement of LLMs and Prompting Approach. The blue line shows the average internal agreement across all papers while the grey lines show the individual results of each paper.}
    \label{fig:internal}
\end{figure}

\subsection{LLM and Manual Consensus Agreement}
\begin{wrapfigure}[21]{r}{0.55\textwidth} 
    \centering
    \includegraphics[width=\linewidth]{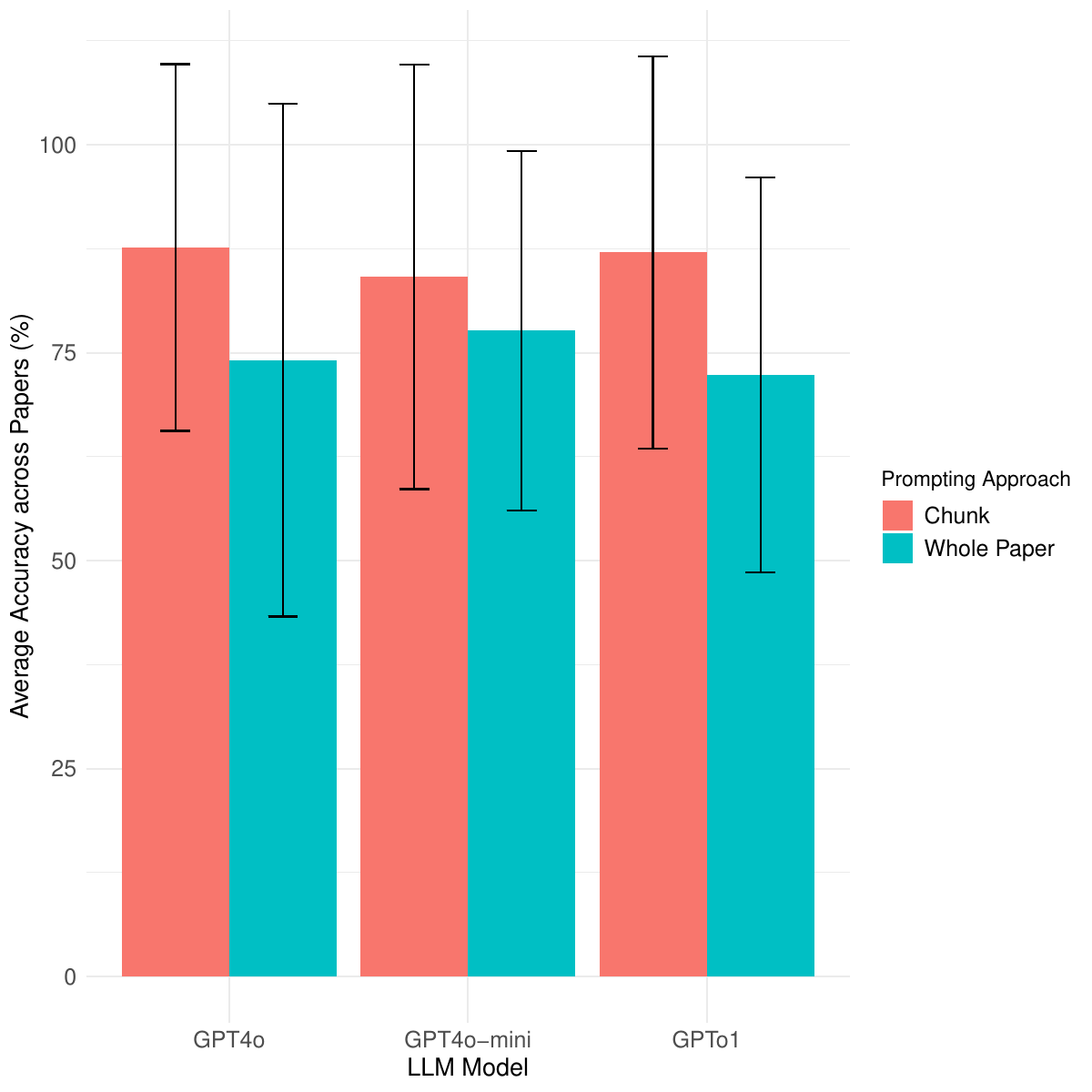}
    \caption{Averaged LLM Accuracy with Manual Raters with $\pm 2$ Standard Deviations of Error.}
    \label{fig:accuracy}
\end{wrapfigure}

Figure \ref{fig:accuracy} illustrates accuracy as a measure of performance of the LLM consensus results using the consensus of the three human raters as the comparison benchmark across all dimension classifications. Across all models, the chunking approaches had a higher mean agreement across the 170 total manual ratings, indicating that models are more likely to agree with human raters across iterations. The highest levels of overall agreement of the LLM compared to human raters were all achieved using the chunking approach Kruskal-Wallis test of variance shows moderate statistical significance between models and coding approaches (p \textless 0.10).     

When examining the recall of the different models and prompting approaches using a Mann-Whitney pairwise comparison, there is a statistically significant difference between several of the groups. Notably, both GPT-4o and o1-mini performed significantly better than all whole paper approaches. This indicates that GPT-4o and o1-mini more correctly identified relevant outcomes when they have a smaller volume of data to sift through. No significant differences were found using GPT-4o-mini when applying the whole text approach or the fixed chunk approach. None of the models performed differently from another when using the chunking approach. GPT-4o-mini performed better than o1-mini when using the whole paper approach, further indicating that GPT-4o-mini is less volume dependent while o1-mini suffers greater when it has more data.

A Kruskal-Wallis test shows that between all groups there is no significant difference in the precision of the models, indicating that there is not a major trade-off between the higher recall of the chunking approaches and precision. This means that GPT-4o or o1-mini employed with the chunking approach are better at correctly identifying relevant outcomes without overclassifying the dimensions as true. 

\begin{figure}[H]
    \centering
    \includegraphics[width=0.55\linewidth]{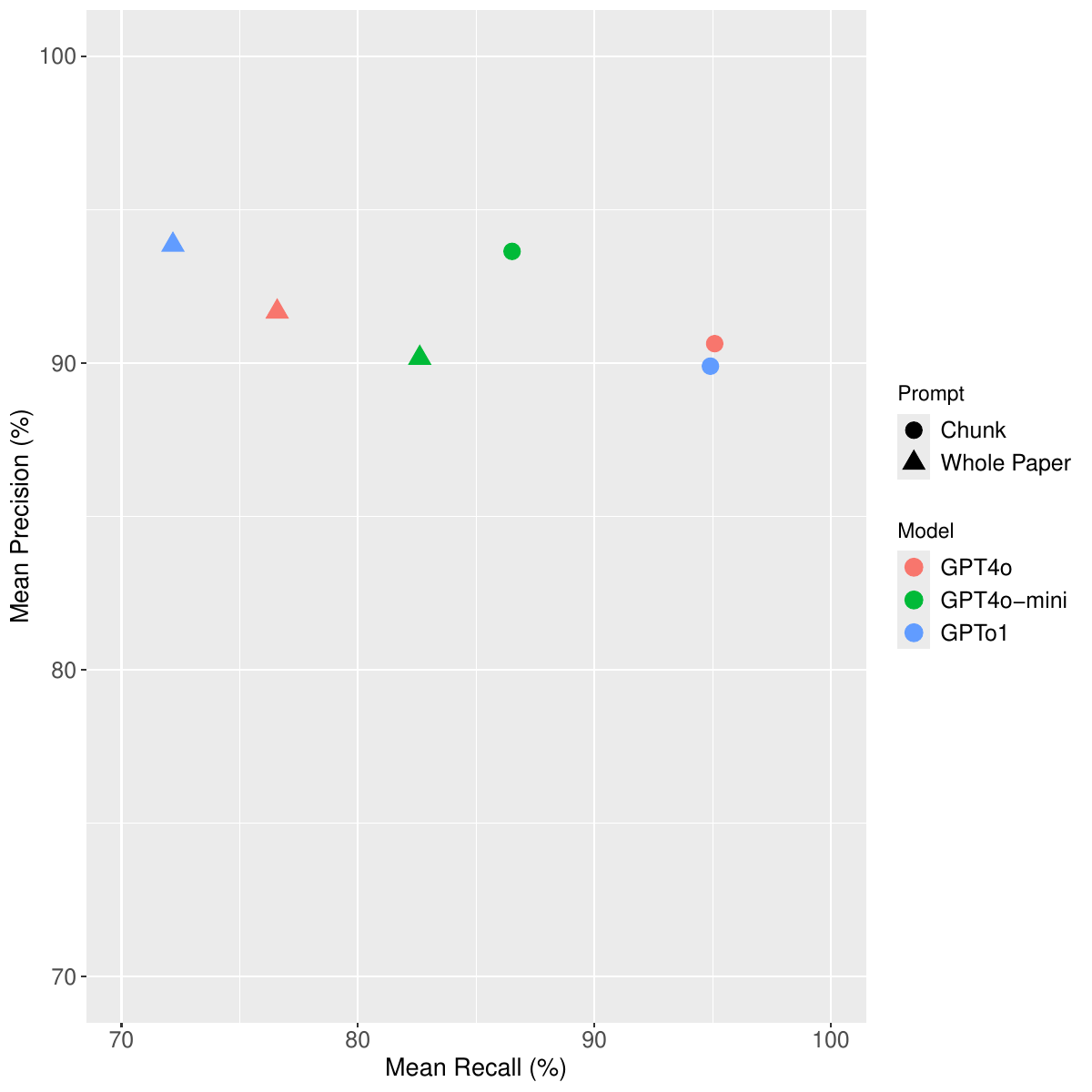}
    \caption{Precision vs. Recall of Models across all classifications using consensus approach when compared to manual consensus results across all 10 analyzed documents.}
    \label{fig:pvt}
\end{figure}

\subsection{Frequency of Classification}
Across all dimensions and papers, the consensus of the manual coders marked 86\% of classifications as True, as displayed in Figure 4. When comparing the positive identification rate (True Positives / Manual Positives), both GPT-4o and o1-mini performed best when employed with the chunking approach. Mann-Whitney pairwise comparisons show these models outperformed the chunking approach with GPT-4o-mini and all whole paper approaches in correctly identifying positives. GPT-4o-mini did not have significant performance difference between the chunking approach and the whole paper approach. This, however, seems to come at the cost of a negative identification rate. As shown above, the chunking approaches for GPT-4o and o1-mini were less likely to classify negatives, resulting in a negative identification rate of 48\% and 45\%, respectively; these are the lowest negative identification rates across the models and prompting approaches. Despite this, a Kruskal-Wallis test shows no significant difference between negative identification rate (True Negatives / Manual Negatives) of the treatment groups, indicating that the decreased negative identification rate of GPT-4o and o1-mini using chunked text is not a statistically significant underperformance.  

\begin{figure}[H]
    \centering
    \includegraphics[width=.75\linewidth]{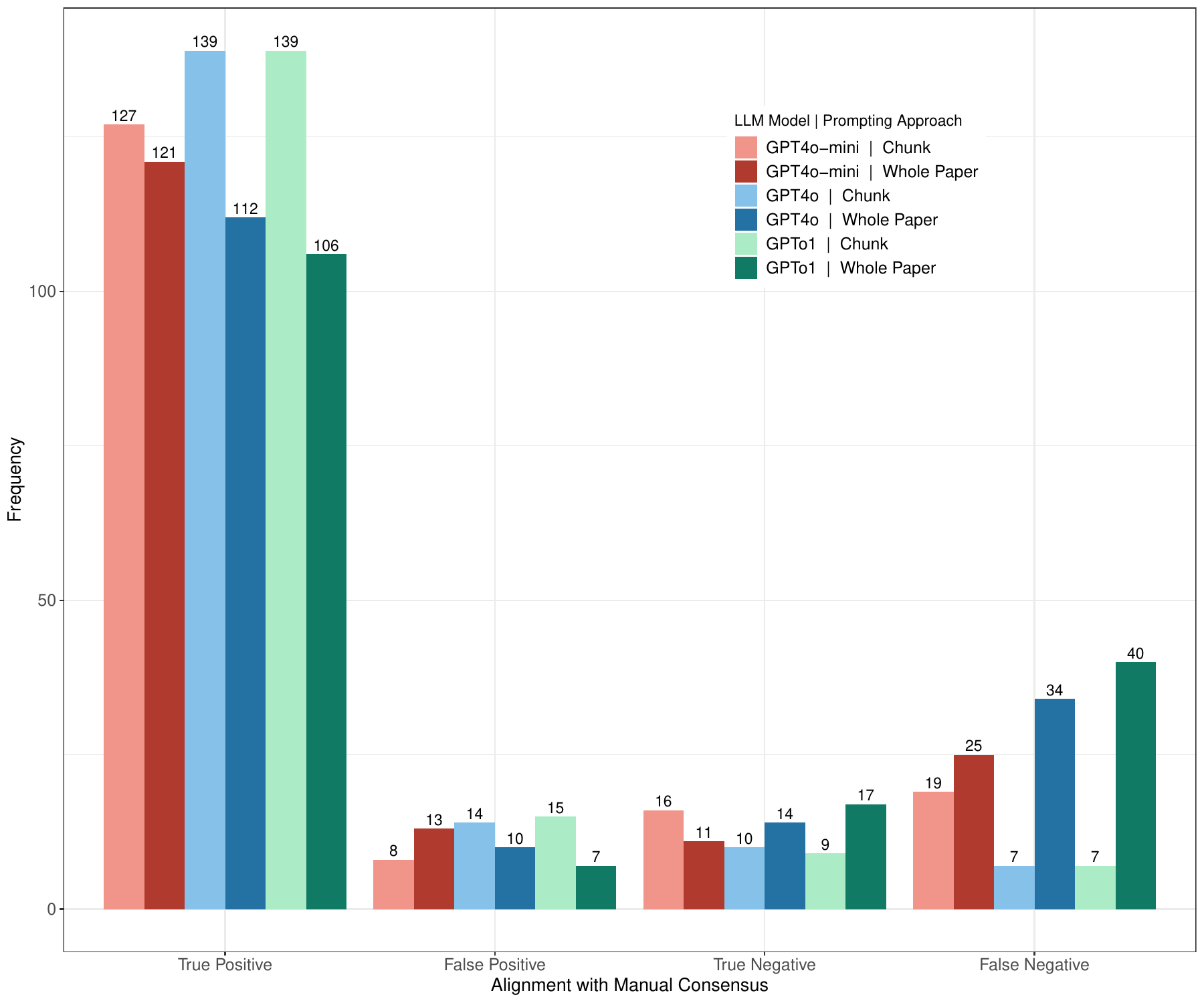}
    \caption{Confusion Matrix results of the different approaches when comparing the consensus of 15 iterations of LLM execution to the consensus of 3 manual raters. }
    \label{fig:cm}
\end{figure}

There were discrepancies within specific coded dimensions towards either overclassifying positives or negatives. Notably, all models and prompting approaches suffered in correctly identifying true instances of the dimensions\textit{ Data Ownership} and \textit{Virtual-to-Physical Connection (V2P)}. Interestingly, manual raters also showed disagreement, with 40\% of papers and 60\% of papers indicating disagreement in the classifications of \textit{Data Ownership} and \textit{V2P}, respectively. The Whole Paper approaches were less accurate in correctly identifying positive instances of the dimensions \textit{Metrology}, \textit{Realization}, and \textit{Virtual Environment} while the chunking approach for all models had higher positive identification rates. The dimension Realization also caused disagreement to manual raters where 60\% of papers had disagreement between raters regarding the discussion of this dimension. \textit{Metrology} and \textit{Virtual Environment} showed 20\% of papers having disagreements between manual raters regarding the discussion of the parameters. 

Failure to identify the lack of discussion regarding a dimension was common for some dimensions as well. All models and prompting approaches struggled to identify the absence of discussion of the dimensions \textit{State} and \textit{Fidelity}. Manual raters disagreed for these dimensions on 30\% and 40\% of papers, respectively. 

\begin{figure} [H]
    \centering
    \includegraphics[width=1\linewidth]{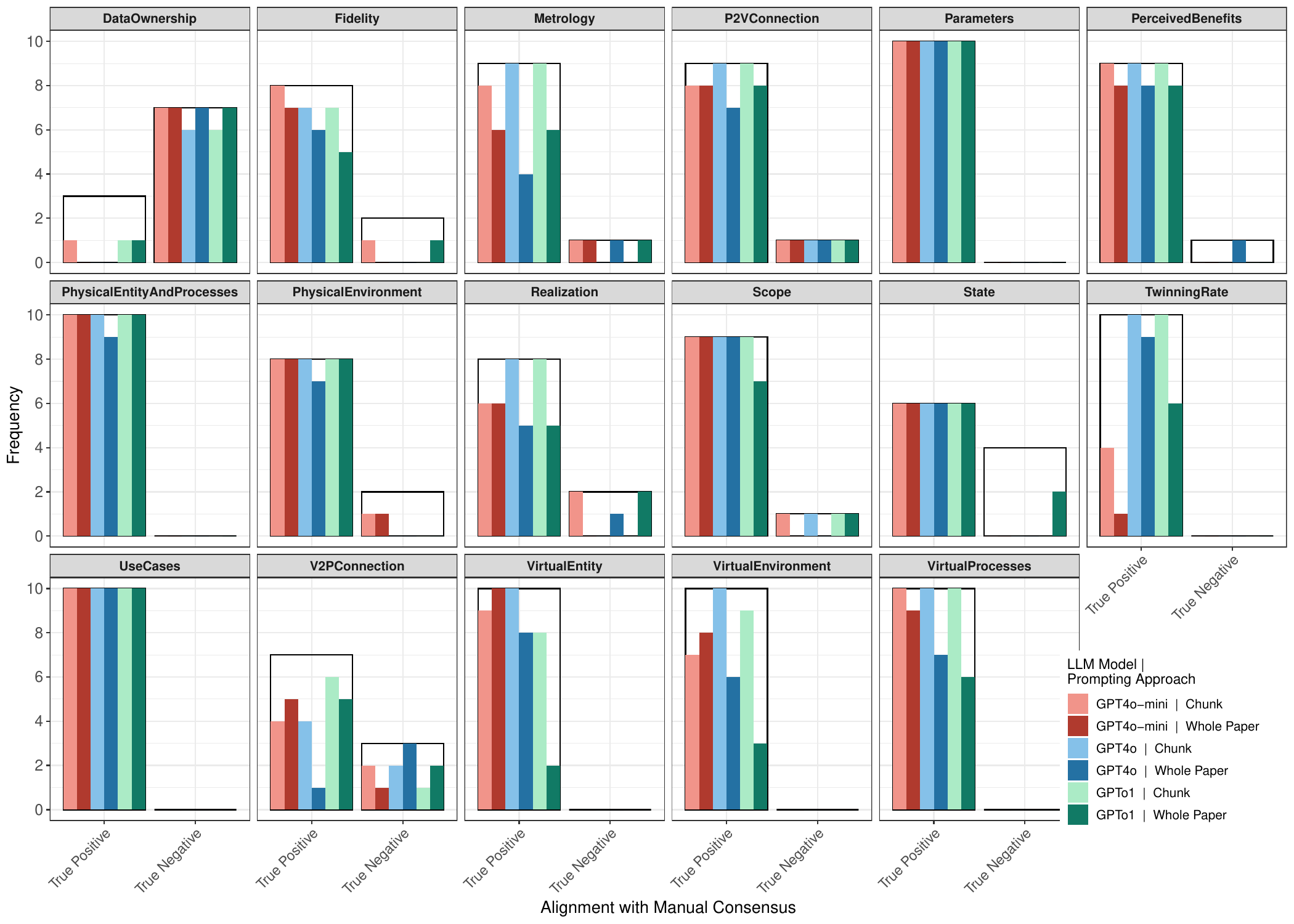}
    \caption{Consensus Approach Comparison to human raters for each dimension across all 10 papers, showing the true positive and true negative classifications. The background white bar shows the number of positive or negative classifications by the manual consensus. }
    \label{fig:cm_dim}
\end{figure}

The manual coding achieved a Fleiss Kappa value of 0.434, indicating fair agreement amongst manual raters, a statistically significant difference from purely chance agreement. As shown in Table 4, the Fleiss K values of the models when employed as a 4th rater decreased across all models when using the whole paper approach. The Chunking approach, on the other hand, did maintain statistical significance when employed with the GPT-4o and GPT-4o-mini models, although only the GPT-4o-mini showed improvements in the K value from the manual baseline. This indicates that the chunking approach is statistically more likely to have non-chance agreements with humans.

\begin{table}[h]
    \centering
    \begin{tabular}{lcccccc}
        \toprule
        \textbf{LLM} & \multicolumn{2}{c}{\textbf{GPT-4o-mini}} & \multicolumn{2}{c}{\textbf{GPT-4o}} & \multicolumn{2}{c}{\textbf{o1-mini}} \\
        \cmidrule(lr){2-3} \cmidrule(lr){4-5} \cmidrule(lr){6-7}
        \textbf{Prompting Approach} & Whole Paper & Chunk & Whole Paper & Chunk & Whole Paper & Chunk \\
        \midrule
        Consensus Kappa (N=4)  & 0.346 & 0.437 & 0.324 & 0.407 & 0.337 & 0.390 \\
        Internal Kappa (N=15)   & 0.942 & 0.644 & 0.911 & 0.596 & 0.663 & 0.634 \\
        \bottomrule
    \end{tabular}
    \caption{Fleiss K values of Models with manual ratings using consensus approach (n=4) and the individual iterations as raters without humans (n=15). When using only the manual results, the Fleiss kappa value across the three raters and the ten papers was 0.434.}
    \label{tab:llm_comparison}
\end{table}

\begin{figure} [H]
    \centering
    \includegraphics[width=1\linewidth]{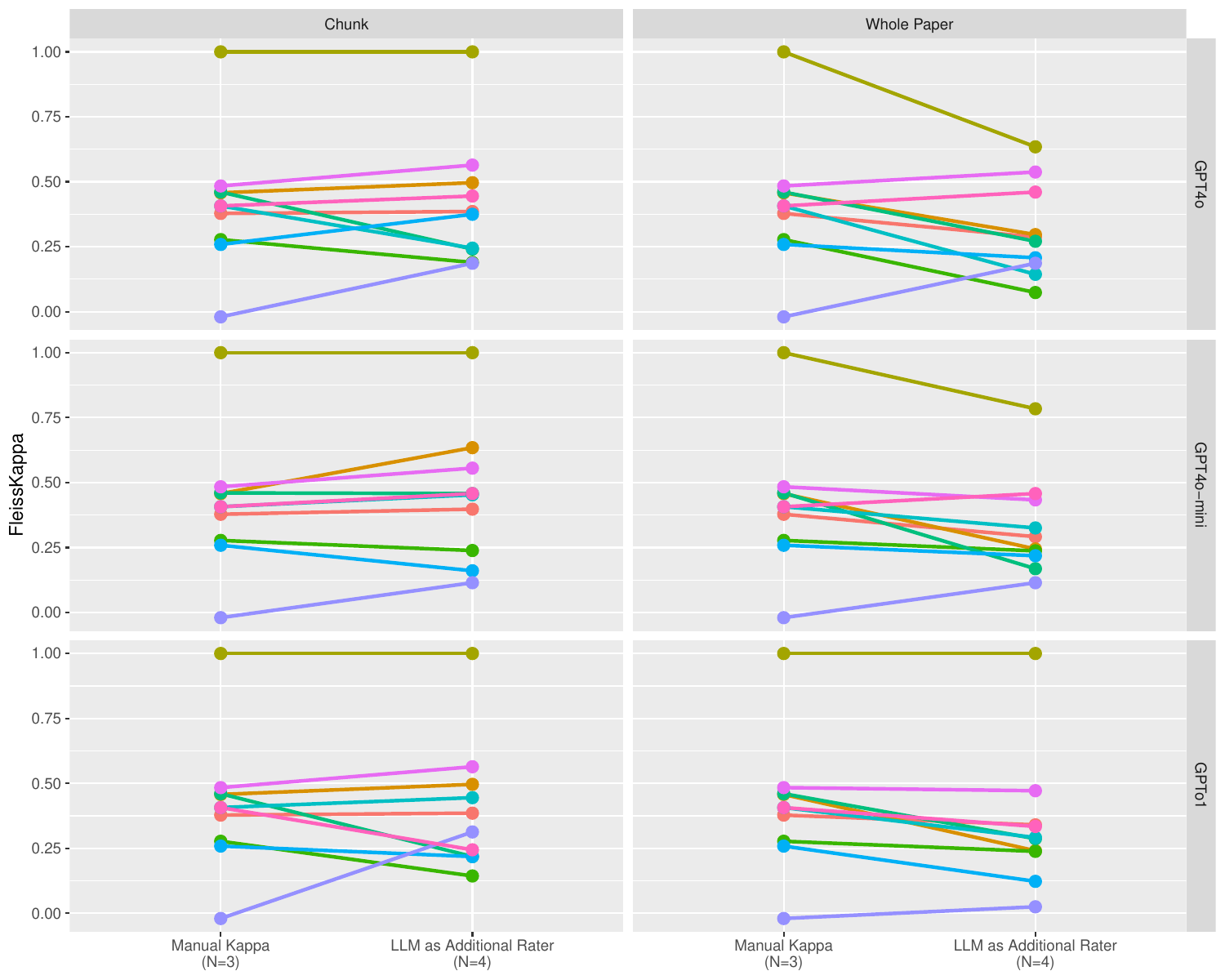}
    \caption{Changes to the Fleiss Kappa interrater agreement statistic by paper when adding the consensus result of the LLMs and prompting approaches as an additional rater.}
    \label{fig:kappa}
\end{figure}

Figure \ref{fig:kappa} shows Kappa statistic after adding the LLM as an additional rater. Six of the ten analyzed articles showed some improvement when using the chunking approach with the o1-mini and GPT-4o-mini models while seven showed some improvement with the GPT-4o model. The frequency of improved analyses dropped precipitously when employing the whole text approach. Employing the chunking approach with both GPT-4o and GPT-4o-mini showed significant agreement when teamed with human raters as indicated by the kappa value above 0.400. o1-mini showed near significant agreement with humans, although ultimately it fell below the threshold. This indicates that, if employed within a human-AI team, the chunking approach for GPT-4o and GPT-4o-mini could consistently improve the significance of the analysis of documents.

\section{Discussion}

This study aimed to provide a comparative exploration of deductive coding methods utilizing LLMs to address the time and variability constraints of human coding of extensive textual documentation. We found that in the context of coding digital twin characteristics in scientific documents relating to urban systems that LLMs have demonstrated potential to address these challenges and provide a viable additional coder. Thereby providing an option for evaluating large datasets and adding additional context for managing complex urban environments. 

This research adopts AI-teaming with deductive coding, leveraging AI's efficiency alongside a coding framework derived from expert knowledge. Our methodology was designed to approximate the rigor of conventional complex binary documentation analysis with the augmented capacities of artificial intelligence. In deploying LLMs, we aimed to refine the coding process specifically within the domain of Urban systems applications. This application of LLMs is intended to augment the speed and analytical precision with which complex scientific papers are classified and examined and thereby aid in the sustainable management of urban systems infrastructure. The findings suggest that when using text chunking strategies, GPT-4o and o1-mini coding results closely align with human coding and perform better than their whole paper counterparts across almost all metrics analyzed in this paper while maintaining internal consistency. This is shown in the significance of the K values shown in Table 4, where both GPT-4o and GPT-4o-mini maintained statistical significance when paired with human raters. These results suggest that LLMs can perform comparably to human coders or assist in analysis when provided specific prompts and tasks structured within the deductive coding process. Utilizing LLMs as an additional coder could not only reduce the time and resource constraints associated with manual coding but provide a more reliable source of coding as once the LLMs are trained it produces compatible results. This suggests an application where AI-enabled coding could substantially augment the document review workflow. While this paper did not explore prompt engineering, future research may focus on how LLM performance compares to people when given different codebook to address some of the issues identified for specific dimensions. 

Moreover, our research highlights the need for analyzing the semantic processing of LLMs. We found that chunking more closely resembles how humans process language. We however caution the readers on blanket application of the findings from this study. The performance demonstrated was specific to the codebook and prompting approach. Further research is needed to consider and address the range of consensus found and research would be beneficial to explore the specifics of how each GPT model explained its findings. This could hopefully lead to methods which can help address failures in the codebook used for deductive coding, reducing uncertainty for humans and AI alike. With greater understanding into the semantic processing of LLMs, prompts and deductive coding codebooks can be engineered for better LLM performance and explainability. Whether these observed effects continue with later versions of the OpenAI or other developed LLMs is an area of potential research. Ultimately, we found that, given the high percentage of false negatives, that certain contextual scanning could lead to anomalous investments. So, while LLMs present a viable tool, their application should be deployed with rigorous validation against human benchmarks. However, one limitation of our approach is that this paper employed fixed-size text chunking based on the number of words in an article which runs the risk of cutting off context at the chunk boundaries. Further research may explore how different chunking approaches impact performance of LLMs, such as semantic chunking or thematic chunking. 

Another notable aspect of our findings relates to the choice of LLMs. Unlike GPT-4o and GPT-4o-mini, which rely on direct-shot responses, o1-mini employs a chain-of-thought approach that appears to confer a unique advantage when dealing with smaller segments of text. Although o1-mini did not outperform GPT-4o under whole-text conditions, its performance generally improved the results of both GPT-4o and GPT-4o-mini when utilizing the chunks approach. This observation aligns with the reasoning patterns reported in recent work on the o1-model series \cite{wang_planning_2024}, which highlight the models’ ability to plan, self-refine, and address tasks incrementally. The recursive thought chain mechanism within o1-mini likely enables it to deeply consider each segment, scanning and synthesizing information in a manner more akin to a human’s incremental reading process. By contrast, GPT-4o and GPT-4o-mini—lacking this recursive reasoning structure—may struggle to maintain comprehensive context or revisit earlier content consistently, particularly as document length expands. Hence, o1-mini’s strong performance in the chunk-based analysis emphasizes how modeling approaches that incorporate structured reasoning steps can yield richer, more reliable qualitative assessments.

\section{Conclusion}

As the volume of textual data used within the urban systems management and development industry ever increases, methods to categorize and link documents are increasingly needed to reduce the cognitive burden on humans. This paper analyzed how large language models could be employed within human-AI teams for the deductive coding and classification of articles, examining their similarities to human raters when employing different prompting approaches. Three models from OpenAI were analyzed - GPT-4o, GPT-4o-mini, and o1-mini – using both a whole text prompting approach and a fixed-size text chunking approach. These models were run for 15 iterations across 10 papers for the analysis of 17 different binary dimensions with a team of 3 non-expert human raters as a point of comparison. For o1-mini and GPT-4o, providing a reduced volume of data for a given execution through text chunking improved the internal agreement across iterations compared to using the same models with the whole text. These models also showed improvements in the recall of data without a significant sacrifice to the precision when employed with text chunking, indicating that chunking aids the models in correctly identifying relevant outcomes; this is also shown through the higher identification of true positives. Finally, GPT-4o, o1-mini and GPT-4o-mini showed significant agreement with human raters when employed using the chunking method, although this agreement rate dropped when passing the whole text for GPT-4o and o1-mini. The application of both GPT-4o and GPT-4o-mini as an additional rater with three manual raters showed statistical significance, indicating that the LLMs can benefit the analysis of textual documents. GPT-4o-mini showed little difference based on the volume of text provided during prompting. In fact, it outperformed o1-mini in the whole text analysis, indicating that o1-mini is greatly impacted by the volume of data passed to the LLM whereas GPT-4o-mini showed minimal variation in performance. Ultimately, this research shows that a fixed-size chunking approach for deductive coding using LLMs can very closely mirror the same analysis performed by humans which can be useful for either augmenting or automating the document classification process. 

\setcounter{table}{0}
\renewcommand{\thetable}{S\arabic{table}}

\subsection*{Funding Statement}
This work was supported through related research funded by the National Alliance for Water Innovation and US Department of Energy (NAWI 3.25), Prof. Steven A Conrad is the corresponding author of the paper.
\subsection*{Acknowledgments}
We are grateful for the support of Colorado State University’s BlueGreen Decisions Lab in the analysis of textual data.

\subsection*{Data Availability}
Supplementary materials, replication data and code can be found on GitHub: \href{https://github.com/josh-rodriguez-csu/ChunkingforLLMs}{https://github.com/josh-rodriguez-csu/ChunkingforLLMs}
\subsection*{Author Contributions}
Author contributions. Conceptualization: O.S. Methodology: O.S.; J.R. Data curation: J.R. Data visualisation: J.R. Writing original draft: J.R.; G.V.L.; O.S.; S.A.C. All authors approved the final submitted draft.


\bibliography{main}

\newpage

\section*{Appendix: Sample LLM Results}
\begin{table}[H]
    \centering
    \small
    \renewcommand{\arraystretch}{1.2} 
    \begin{tabular}{p{2.8cm} p{6.5cm} p{6.5cm}}
        \toprule
        \textbf{Model} 
        & \multicolumn{2}{c}{\textbf{Classification Type}} \\
        \cmidrule(lr){2-3}
         & \textbf{Positive} & \textbf{Negative} \\
        \midrule
        
        \textbf{GPT-4o-mini} &
        Yes, the parameter 'Virtual Entity' is discussed in the text, particularly in relation to the concept of digital twins. Evidence of this can be found in the section where the authors define a digital twin as a "virtual twin" that serves as a digital representation of an object, system, or process. They clarify that digital twins are updated from near-real-time data and assist in decision-making through simulation and analysis ("Digital Twins. A digital twin, sometimes called a virtual twin, is a digital representation of an object, system, or process that simulates real-world performance."). &
        The parameter "Physical Environment" is not directly mentioned or discussed in the provided text. While the text revolves around technological advancements, data management, and innovations in the water utility sector, it primarily emphasizes aspects related to digital transformation rather than the physical surroundings and infrastructure where these systems operate. \\
        
        \midrule
        \textbf{GPT-4o} &
        The text discusses 'Physical Entity and Processes' by referencing tangible objects or real-world processes within physical systems. Evidence of this is seen in the discussion about technologies such as digital twins and AI, which help visualize, analyze, and optimize complex systems like treatment plants (paragraph 11). Additionally, the text describes how real-time control systems, such as SCADA, are used to manage water distribution systems, which are operational processes in industrial settings (paragraph 21). These examples highlight the involvement of tangible equipment and systems in optimizing industrial operations, fitting the definition of 'Physical Entity and Processes.' &
        The text does not specifically address 'Data Ownership,' as defined as 'The legal ownership and security of the data stored within the Digital Twin.' Instead, the article focuses on the applications and transformative potential of digital technologies like artificial intelligence, digital twins, and data integration in the water utility sector. The text emphasizes the ability of these technologies to enhance operational efficiency, improve decision-making, and foster collaboration by using data-driven insights but does not delve into the legal or security aspects of data ownership. \\
        
        \midrule
        \textbf{o1-mini} &
        Yes, the parameter 'Use Cases' is discussed in the text. The article provides specific applications of digital twins, such as automating public notifications for combined system overflows to ensure regulatory compliance and protect public health. Additionally, it highlights the use of AI for predictive maintenance, enabling utilities to forecast equipment failures and optimize maintenance schedules. These examples demonstrate how digital twins and AI serve as decision support and forecasting tools within water utility operations, aligning with the defined concept of 'Use Cases.' &
        The term "Virtual Entity" is not explicitly mentioned in the provided text. While the article extensively discusses \textbf{digital twins} and \textbf{virtual twins}, which are related concepts, it does not specifically address individual digital counterparts or replicas as defined by "Virtual Entity." For example, the text states, "Digital twins and AI enable users to visualize, analyze, and optimize complex systems," and further elaborates on the functionalities of digital twins. However, it does not delve into the specifics of digital models mimicking individual physical entities within the digital twin environment. \\
        
        \bottomrule
    \end{tabular}
    \caption{Sample Positive and Negative Results from each LLM for a given Dimension.}
    \label{tab:sample}
\end{table}

\end{document}